\documentclass[preprint,12pt,3p]{elsarticle}

\usepackage{amssymb}
\usepackage{graphicx}
\usepackage{amsmath}

\usepackage{arydshln}

\usepackage{subcaption}
\usepackage{slashbox}
\usepackage[export]{adjustbox}

\newcommand{\cN}{{\cal N}}

\newcommand{\cR}{{\cal R}}

\newcommand{\cX}{{\cal X}}
\newcommand{\cY}{{\cal Y}}

\makeatletter
\def\ps@pprintTitle{%
	\let\@oddhead\@empty
	\let\@evenhead\@empty
	\def\@oddfoot{\reset@font\hfil\thepage\hfil}
	\let\@evenfoot\@oddfoot
}
\makeatother

\usepackage[hyphens]{url}
\begin{document}

\begin{frontmatter}

\title{Noisy Computations during Inference: Harmful or Helpful?}

\author{\corref{cor1}Minghai Qin}

\cortext[cor1]{I am corresponding author}

\ead{qinminghai@gmail.com}

\author{Dejan Vucinic}


\begin{abstract}
We study two aspects of noisy computations during inference. 
The first aspect is how to mitigate their side effects for naturally trained deep learning systems. 
One of the motivations for looking into this problem is to reduce the high power cost of conventional computing of neural networks through the use of analog neuromorphic circuits. 
Traditional GPU/CPU-centered deep learning architectures exhibit bottlenecks in power-restricted applications (e.g., embedded systems). 
The use of specialized neuromorphic circuits, where analog signals passed through memory-cell arrays are sensed to accomplish matrix-vector multiplications, promises large power savings and speed gains but brings with it the problems of limited precision of computations and unavoidable analog noise. 
We manage to improve inference accuracy from $21.1\%$ to $99.5\%$ for MNIST images, from $29.9\%$ to $89.1\%$ for CIFAR10, and from $15.5\%$ to $89.6\%$ for MNIST stroke sequences with the presence of strong noise (with signal-to-noise power ratio being 0 dB)  by noise-injected training and a voting method.
This observation promises neural networks that are insensitive to inference noise, which reduces the quality requirements on neuromorphic circuits and is crucial for their practical usage. 

The second aspect is how to utilize the noisy inference as a defensive architecture against black-box adversarial attacks. During inference, by injecting proper noise to signals in the neural networks,  the robustness of adversarially-trained neural networks against black-box attacks has been further enhanced by $0.5\%$ and $1.13\%$ for two adversarially trained models for MNIST and CIFAR10, respectively. 
\end{abstract}


\end{frontmatter}


\section{Introduction}

Neural networks (NNs)~\cite{Schmidhuber15,HTF01} are layered computational networks that try to imitate the function of neurons in a human brain during object recognition, decision making, and other cognitive tasks. They are one of the most widely used machine learning techniques due to their good performance in practice. Certain variants of neural networks were shown to be more suitable for different learning applications. For instance, deep convolutional neural networks (CNNs)~\cite{JKR09,KSH12} were found to be effective at recognizing and classifying images. Recurrent neural networks (RNNs)~\cite{GLF09,HAF14} provide stronger performance at sequence predictions, e.g. speech or text recognition. A neural network is determined by the connections between the neurons and weights/biases associated with them which are trained using the back-propagation algorithm~\cite{RB93,Nielsen89}.

In order to fit highly non-linear functions and thus to achieve a high rate of correctness in practice, neural networks usually contain many layers, each containing a large number of weights. The most power- and time-consuming computation of a neural networks is the matrix-vector multiplication between weights and incoming signals. 
In power-restricted applications, such as an inference engine in an embedded system, the size of the neural networks is effectively limited by the power consumption of the computations.
One attractive method of lowering this power consumption is by use of neuromorphic computing~\cite{ZADFBG10}, where analog signals passed through memory-cell arrays are sensed to accomplish matrix-vector multiplications. Here weights are programmed as conductances (reciprocal of resistivity) of memory cells in a two-dimensional array, such as resistive RAM (ReRAM), phase-change RAM (PCM) or NAND Flash~\cite{MME13}. According to Ohm's law, if input signals are presented as voltages to this layer of memory cells, the output current is the matrix-vector multiplication of the weights and the input signals. The massive parallelism of this operation also promises significant savings in latency over sequential digital computations.

One of the problems with neuromorphic computing is the limited precision of the computations, since the programming of memory cells, as well as the measurement of the analog outputs, inevitably suffers from analog noise, which affects performance of deep learning systems in a harmful way.
Some researchers have proposed to correct the errors between A/D and D/A converters of each layer~\cite{FWI18,JRRR17}, which induces extra overhead in latency, space, and power as well. Existing noisy inference literature for neuromorphic computing is mostly in the device/circuit domain, typically focusing on redesigning memory devices/circuits to reduce the noise power. 
But to our knowledge,  the exploration of robustness of neural networks against analog and noisy computations inside the network in an algorithmic point of view has not been studied in detail. It is the first goal of this paper to present methods and observations to mitigate the harmful effects brought by inference noise.

During its study, we find out that noisy inference can also be helpful to defend against black-box adversarial attacks of neural networks, the resistance of which is becoming more crucial for a wide range of applications, such as deep learning systems deployed in financial institutions and autonomous driving. 
Adversarial attacks manipulate inputs to a neural networks and generate adversarial examples that have rarely perceptible differences to a natural image, but they lead to incorrect predictions~\cite{Szegedy2013IntriguingPO}. 
White-box attacks can be categorized into three categories. 
\begin{itemize}
	\item Single step attack, e.g., fast gradient sign methods (FGSM)~\cite{GoodfellowExplaining14},
	\[
	x^* = x + \epsilon \cdot   \textrm{sign} (\nabla_x J(x,y)),
	\] 
	where $x$ and $x^*$ is the natural and adversarial examples, respectively. $\epsilon$ is the upper bound on the $L_{\infty}$ distance between $x$ and $x^*$. $J(x,y)$ is the loss function for an input $x$ being predicted to have label $y$ and $\nabla_x$ is the derivative operator with respect to input $x$.
	
	\item Iterative methods, e.g., iterative FGSM~\cite{KurakinAdversarialPhysical16},
	\[
	x^*_0 = x, x_{t+1}^* = \textrm{Clip}_{x,\epsilon} \left(x_{t}^* + \alpha \cdot \textrm{sign} (\nabla_x J(x_t^*,y)) \right),
	\]
	where $\alpha$ is the learning rate of the adversary, Clip$_{x,\epsilon}(\cdot)$ is a function that clips the output to be within the $\epsilon$-cube of $x$ such that the $L_{\infty}$ distance between $x^*_t$ and $x$ is always less than or equal to $\epsilon$.
	
	\item Optimization based attacks, e.g.,~\cite{CarliniEvaluatingRobustness16},
	\[
	\arg \min_{x^*} \lambda || x - x^*||_p + J(x^*,y),
	\]
	where $\lambda$ balances the $L_p$ norm distortion and the loss function.
\end{itemize}

Black-box attacks are possible based on the transferability~\cite{PapernotTransferability16} of adversary such that adversarial examples generated from one neural network often lead to incorrect predictions of another independently trained neural network.

Several defensive mechanisms such as distillations~\cite{PapernotDistill15}, feature squeezing~\cite{XuFeatureSqueezing17}, and adversary detection~\cite{Feinman2017DetectingAS}, are proposed against attacks.~\cite{madry2018towards} models the process as a min-max problem and builds a universal framework for attacking and defensive schemes, where it tries to provide a guaranteed performance for the first-order attacks, i.e., attacks based soly on derivatives. 
They also suggested that adversarial training with projected gradient descent (PGD) provides a strongest robustness against first-order attacks.
Based on their study, we further investigate noisy inference of adversarially trained  neural networks. Experimental results show improvement on robustness against black-box adversarial attacks and indicate that adversarial examples are more sensitive to inference noise than natural examples.

The contribution of this paper is summarized as follows.

1. We model the analog noise of neuromorphic circuits as additive and multiplicative Gaussian noise. The impact on accuracy of noisy inference is shown to be severe by three neural network architectures (a regular CNN,  a DenseNet~\cite{HLW16}, and a LSTM-based RNN) for three datasets (MNIST, CIFAR10, and MNIST stroke sequences~\cite{JongMnistStroke2016}), e.g., when noise power equals the signal power, the accuracy is as low as $(21.1\%, 29.9\%, 10.1\%)$, respectively.

2. We observe that the performance of noisy inference can be greatly improved by injecting noises to all layers in the neural networks  during training. Noise-injected training was used as a regularization tool for better model generalization, but its application to noisy inference has not been studied in details. We provide a quantitative measurement on the impact that the inference noise has on noiseless-trained and noise-injected trained neural networks, where the power of training and inference noise might not match. The performance of noisy inference with low-to-medium noise power is improved to almost as good as noiseless inference. For large noise power (equal to signal power), the accuracy is increased to $(77.7\%, 66.9\%, 66.9\%)$ for the three datasets, respectively.

3. We further improve the performance of noisy inference by proposing a voting mechanism for large noise power. The accuracy has been further increased to $(99.5\%, 89.1\%, 89.6\%)$ for the three datasets when noise power equals the signal power. In addition, we observe that with noise-injected training and the proposed voting mechanism combined, noisy inference can give higher accuracy $(0.5\%)$ than noiseless inference for LSTM-based recurrent neural networks, which is counterintuitive since it is often believed that noise during inference would be harmful rather than helpful to the accuracy.

4. A further study on adversarially trained neural networks for MNIST and CIFAR10 has shown that noisy inference improves the robustness of deep neural networks against black-box attacks. The accuracy of adversarial examples has been improved by $0.5\%$ and $1.13\%$ (in absolute values) for MNIST and CIFAR10 when validated on a separately and adversarially trained CNN and DenseNet, respectively.

\section{Preliminaries}\label{sec:prelim}
A neural network contains input neurons, hidden neurons, and output neurons. It can be viewed as a function $f:\cX \rightarrow \cY$ where the input $x\in\cX\subseteq\cR^n$ is an $n$-dimensional vector and the output $y\in\cY\subseteq\cR^m$ is an $m$-dimensional vector. In this paper, we focus on classification problems where the output $y=(y_1,\ldots,y_m)$ is usually normalized such that $\sum_{i=1}^m y_i = 1$ and $y_i$ can be viewed as the probability for some input $x$ to be categorized as the $i$-th class. The normalization is often done by the softmax function that maps an arbitrary $m$-dimensional vector $\hat y$ into normalized $y$, denoted by $y=softmax(\hat y)$, as $y_i = \frac{\exp(\hat y_i)}{\sum_{i=1}^m \exp(\hat y_i)}, i = 1,\ldots,m$. For top-$k$ decision problems, we return the top $k$ categories with the largest output $y_i$. In particular for hard decision problems where $k=1$, the classification results is then $\arg\max_i y_i, i=1,\ldots,m$.

A feed-forward neural network $f$ that contains $n$ layers (excluding the softmax output layer) can be expressed as a concatenation of $n$ functions $f_i, i = 1,2,\ldots,n$ such that $f = f_n(f_{n-1}(\cdots f_1(x)\cdots))$. The $i$th layer $f_i: \cX_i\rightarrow \cY_i$ satisfies $\cY_i\subseteq \cX_{i+1}$, $\cX_1 = \cX$. The output of last layer $\cY_n$ is then fed into the softmax function. The function $f_i$ is usually defined as
\begin{align}\label{eq:wx+b}
f_i(x) = g(W \cdot x + b), 
\end{align}
where $W$ is the weights matrix, $b$ is the bias vector, and $g(\cdot)$ is an element-wise activation function that is usually nonlinear, e.g., tanh, sigmoid,  rectified linear unit (ReLU)~\cite{HSMDS00} and leaky ReLU~\cite{MHN13}. Both $W$ and $b$ are trainable parameters.

In this paper, two noise models are assumed, which are additive Gaussian noise model $z\sim\cN(0,\sigma^2)$ and multiplicative Gaussian noise model  $z\sim\cN(1,\sigma^2)$, respectively, in the forward pass after each matrix-vector multiplication. Then Eq.~$(\ref{eq:wx+b})$ becomes
\begin{align}\label{eq:wx+b+z}
f_i(x) = g(W \cdot x + b + z), 
\end{align}
for additive Gaussian noise or
\begin{align}\label{eq:wx+b*z}
f_i(x) = g( (W \cdot x + b) \cdot z)
\end{align}
for multiplicative Gaussian noise.
Additive Gaussian noise $z\sim\cN(0,\sigma^2)$ models procedures of  neuromorphic computing where the noise power is irrelevant to the signals, such as signal sensing, memory reading, and some random electrical perturbation of circuits. On the other hand for multiplicative noises,  we can show that multiplying a unit-mean Gaussian random variable $Z$ to a signal $X$ is equivalent to adding a zero-mean Gaussian random variable $Z'$ to the signal where the standard deviation of $Z'$ is proportional to the magnitude of the signal, as
\begin{align}
XZ = X(1+Z') = X + XZ', 
\end{align}
where $XZ'~\sim \cN(0,\sigma^2X^2)=\cN(0, (|X|\sigma)^2)$.
Therefore, multiplicative noise models the procedures where the noise power is proportional to the signal power, such as memory programming and computations.

For both additive and multiplicative Gaussian noise models, we denote $\sigma_{\textrm{train}}$ and $\sigma_{\textrm{val}}$ the standard deviation of the noise during training and validation (inference), respectively. We denote $\sigma_{\textrm{train},+}$, $\sigma_{\textrm{val},+}$, $\sigma_{\textrm{train},\times}$, $\sigma_{\textrm{val},\times}$ if specific additive or multiplicative noise models are referred to.
Note that $\sigma_{\textrm{train}}$ is usually set to be $0$ in conventional training; Nonetheless, random noise is sometimes injected ($\sigma_{\textrm{train}}>0$) during training to provide better generalization. Conventional deep learning architecture assumes $\sigma_{\textrm{val}}=0$ since digital circuits are assumed to have no errors during computations.

For both noisy models, the signal-to-noise power ratio (SNR) is a defining parameter that measures the strength of the noise relative to the signal, which is defined as the ratio between the power of signal and noise. It is usually expressed in dB where SNR(dB) = $10\log_{10}(\textrm{SNR(value)})$. For example, if the signal and noise has the same power, SNR = 1 (or $0$ dB). For multiplicative noise models, we have constant SNR = $\frac{1}{\sigma_{\textrm{val},\times}^2}$; on the other hand, SNRs for additive noise models with fixed $\sigma_{\textrm{val},+}$'s depend highly on signal power and it would be a fair comparison only if the signal power is invariant. Therefore, we mainly use multiplicative noise models in our experiment except for the case where we can normalize the signals to have constant power.

\section{Robustness of NNs against Noisy Inference} \label{sec:cnn}
In this section  we explore the robustness of neural networks against noisy computations modeled in Eq.~$(\ref{eq:wx+b+z})$ and Eq.~$(\ref{eq:wx+b*z})$. We will show the robustness of different neural network architectures for different datasets against noisy inference and provide two techniques that improve the robustness, namely, noise-injected training and voting.

\subsection{Datasets and NN architectures}
Three datasets are used in our experiments, which are MNIST images ($60000$ for training and $10000$ for validation), CIFAR10 images ($50000$ for training, $10000$ for validation, no data augmentation applied), and the stroke sequence of MNIST~\cite{JongMnistStroke2016} for each MNIST image.

For MNIST images, we use a 6-layer convolutional neural network described in Table~\ref{tab:cnn6_mnist}. The noiseless trained model has prediction accuracy of $99.5\%$ for noiseless inference.
For multiplicative Gaussian noise models, the parameters in batch normalization layers are trainable. However, for additive Gaussian noise models, we use batch normalization layers with fixed parameters $\beta=0, \gamma=1$ (as opposed to trainable mean and variance) to minimize the signal power variations for the reason mentioned in Section~\ref{sec:prelim}.

\begin{table}[htbp]
	{
		\begin{center}
			
			\caption{A 6-layer CNN architecture for MNIST}
			\label{tab:cnn6_mnist}
			
			\begin{tabular}{|c:c|}	
				\hline Layer & Output shape \\
				\hdashline Conv. 2D $(3,3)$ - Batch Norm $(\beta,\gamma)$  & $(26,26,32)$ \\
				\hdashline Gaussian Noise - ReLu & $(26,26,32)$ \\
				\hdashline Conv. 2D $(3,3)$ - Batch Norm $(\beta,\gamma)$  & $(24,24,32)$ \\
				\hdashline Gaussian Noise - ReLu & $(24,24,32)$ \\
				\hdashline Maxpooling & $(12,12,32)$ \\
				\hdashline Conv. 2D $(3,3)$ - Batch Norm $(\beta,\gamma)$  & $(10,10,64)$ \\
				\hdashline Gaussian Noise - ReLu & $(10,10,64)$ \\
				\hdashline Conv. 2D $(3,3)$ - Batch Norm $(\beta,\gamma)$  & $(8,8,64)$  \\
				\hdashline Gaussian Noise - ReLu & $(8,8,64)$ \\
				\hdashline Maxpooling & $(4,4,64)$ \\
				\hdashline Fully connected - Batch Norm $(\beta,\gamma)$ & $(256)$ \\
				\hdashline Gaussian Noise - ReLu  & $(256)$ \\
				\hdashline Fully connected -  Gaussian Noise& $(10)$ \\
				\hline 
			\end{tabular} 
		
			%
		\end{center}
	}
\end{table}

For CIFAR10, we use a densely connected convolutional neural network (DenseNet)~\cite{HLW16}. DenseNet is an enhanced version of ResNet~\cite{HeResNet15} where all feature maps of previous layers are presented as the input to later convolutional layers. The depth of the DenseNet is $40$ and growth rate is $12$. All parameters in convolutional, fully-connected and batch normalization layers are trainable since fixed batch normalization parameters cannot provide satisfactory accuracy even if the inference is noiseless. The total number of trainable parameters of the DenseNet is around one million. The noiseless trained model has prediction accuracy of $92.5\%$ with noiseless inference, which is slightly below the reported value ($93\%$) in~\cite{HeResNet15}. We inject multiplicative Gaussian noise after each matrix-vector multiplication.

\begin{figure}
	\centering
	\includegraphics[width=0.7\linewidth]{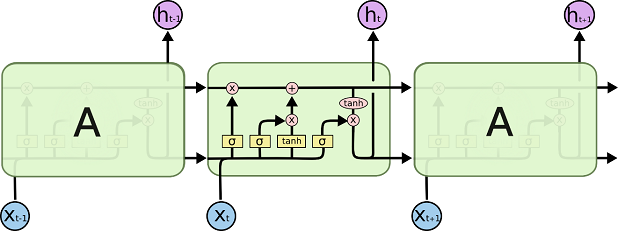}
	\caption{Example of LSTM cells.}
	\label{fig:lstm}
\end{figure}

For MNIST stroke sequences extracted by~\cite{JongMnistStroke2016}, we use a LSTM-based recurrent neural network with $50$ cells corresponding to the first $50$ two dimensional coordinates of pen-points of each written digit. If the total number of pen-points is less than $50$, it is padded with $(0,0)$'s. Since the pen-points are two dimensional, the input dimension to each LSTM cell is $2$. The number of hidden units in each cell is $128$, and the last LSTM cell are connected with an output layer of $10$ neurons for classification. The noiselessly trained model has accuracy around $95\%$ with noiseless inference. The drop of accuracy from MNIST images to MNIST stroke sequences may due to that we have to truncate or pad pen-points to fit the $50$ LSTM cells and gray level information is lost when converting images to stroke sequences.
There are four matrix-vector multiplications in each LSTM cell (see four yellow boxes in LSTM cells in Fig.~\ref{fig:lstm}) and one between the LSTM cell and the output layer, where multiplicative noise are injected. 

\subsection{Noisy Inference by Noiseless and Noise-Injected Training}

In this section, we show the impact of noisy inference on accuracy  for noiselessly and noise-injected trained models. The noise-injected training applies noise layers during training and thus the information in the forward pass is changed and consequently influences the weight updates during backward pass. We train models with different $\sigma_{\textrm{train}}$'s and use different $\sigma_{\textrm{val}}$'s to test the accuracy on all trained models. That is, the noise power in training does not necessarily match the noise power during inference. We will show that such noise-injected training uniformly  improves accuracy on all $\sigma_{\textrm{val}}$'s.

\subsubsection{The CNN for MNIST images}
\begin{figure}[htbp]
	
	\centering
	\includegraphics[width=1\linewidth]{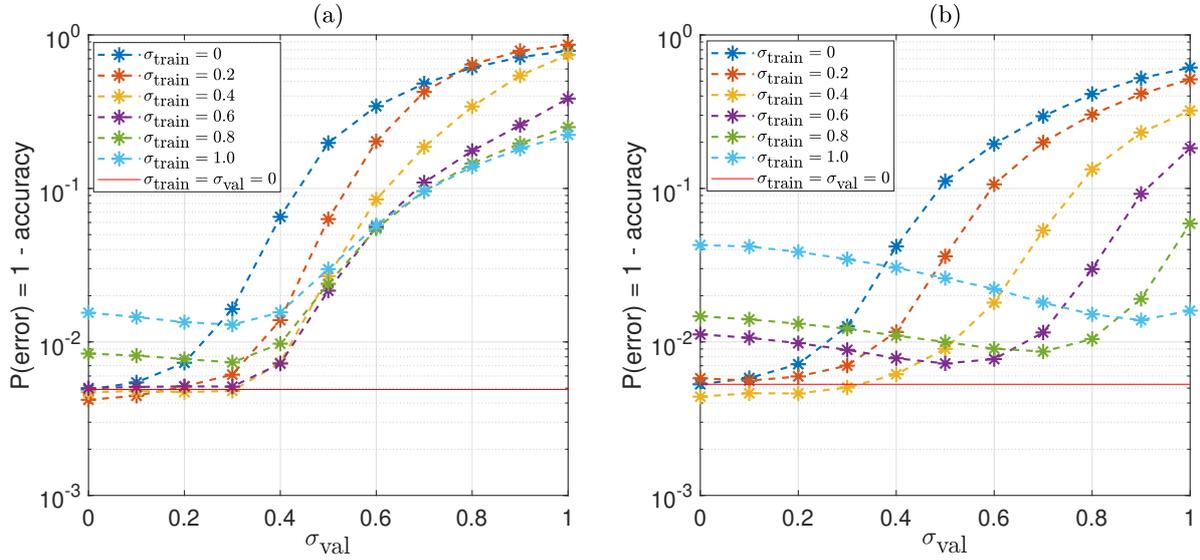}
	\caption{Validation accuracy of the 6-layer CNN with Gaussian noise for different $(\sigma_{\textrm{train}},\sigma_{\textrm{val}})$ pairs. (a) Multiplicative noise model. (b) Additive noise model.}
	\label{fig:cnn6_mnist_mul_add_noisy_training}
\end{figure}

For MNIST images with the 6-layer CNN architecture, we train 100 epochs using stochastic mini-batch training with Adam optimizer and batch size being 32 for each $\sigma_{\textrm{train}}=0.0$ to $1.0$ with a step size of $0.1$. The accuracy of noiselessly trained model $(\sigma_{\textrm{train}}=0)$ with noiseless inference $(\sigma_{\textrm{val}}=0)$ is around $99.5\%$. Then the $11$ trained models are tested on validation set with $\sigma_{\textrm{val}}=0.0$ to $1.0$ with a step size of $0.1$. We present the results of a subset of $\sigma_{\textrm{train}}$'s in Figure~\ref{fig:cnn6_mnist_mul_add_noisy_training}. Validation accuracy for each $(\sigma_{\textrm{train}},\sigma_{\textrm{val}})$ pair is the average of $50$ independent runs and we observe that all $50$ results are highly concentrated in their average (within less than $1\%$ variation). This phenomenon is also confirmed from neural network models for other datasets, thus we use the average accuracy as a measurement of robustness against noisy inference. Also note that $y$-axis represents the error rate ($1-$accuracy) in logarithm scale.
Table~\ref{tab:accu_mul_mnist} and Table~\ref{tab:accu_add_mnist} summarize the results. The first row is the inference noise levels from $\sigma_{\textrm{val}}=0.0$ to $1.0$ with a step size of $0.2$; the second row shows the accuracy for noiselessly trained models; the third row shows the best accuracy for $\sigma_{\textrm{val}}$ in that column, which is achieved by noise-injected training with $\sigma_{\textrm{train}}$ in the fourth row.

\begin{table}[htbp]
	\center
	\caption{Accuracy of noisy inference for multiplicative noise models for MNIST images.}
	\label{tab:accu_mul_mnist}
	{
		\begin{tabular}{|c|c|c|c|c|c|c|}
			\hline 
			{	\scriptsize$\sigma_{\textrm{val},\times}$} & 0.0 & 0.2 & 0.4 & 0.6 & 0.8 & 1.0 \\ 
			\hline 
			{\scriptsize$\sigma_{\textrm{train},\times}$}=0 & 99.5\% & 99.3\% & 93.5\% &65.7\% & 38.8\% & 21.1\% \\ 
			\hline 
			Best accu. & 99.6\% & 99.5\% & 99.3\% & 94.6\% & 86.1\% & 77.7\% \\ 
			\hline 
			By  {\scriptsize$\sigma_{\textrm{train},\times}$} & 0.2 & 0.5 & 0.5 & 0.7 & 1.0 & 1.0 \\ 
			\hline 
		\end{tabular}
	}
	
\end{table}

\begin{table}[htbp]
	\center
	\caption{Accuracy of noisy inference for additive noise models for MNIST images.}
	\label{tab:accu_add_mnist}
	{
		\begin{tabular}{|c|c|c|c|c|c|c|}
			\hline 
			{	\scriptsize$\sigma_{\textrm{val},+}$} & 0.0 & 0.2 & 0.4 & 0.6 & 0.8 & 1.0 \\ 
			\hline 
			{\scriptsize$\sigma_{\textrm{train},+}$}=0 & 99.5\% & 99.3\% & 95.8\% &80.5\% & 58.8\% & 38.7\% \\  
			\hline 
			Best & 99.6\% & 99.5\% & 99.4\% & 99.2\% & 99.0\% & 98.4\% \\ 
			\hline 
			By  {\scriptsize$\sigma_{\textrm{train},+}$} & 0.4 & 0.4 & 0.4 & 0.6 & 0.9 & 1.0 \\ 
			\hline 
		\end{tabular}
	}
	
\end{table}

There are some observations that we can make from Table~\ref{tab:accu_mul_mnist}, Table~\ref{tab:accu_add_mnist}, and Figure~\ref{fig:cnn6_mnist_mul_add_noisy_training}.

1. Conventional noiseless training ($\sigma_{\textrm{train}}=0$) does not provide satisfactory performance against noisy inference even when the noise power is from low to medium (second row in Table~\ref{tab:accu_mul_mnist} and Table~\ref{tab:accu_add_mnist}). For example, when $\sigma_{\textrm{val}}=0.4$, the accuracy reduces by $5.8\%$ and $3.6\%$ for multiplicative and additive noise models, respectively.

2. If the inference noise power is known, training with properly injected noise can greatly improve noisy inference (third and fourth row in Table~\ref{tab:accu_mul_mnist} and Table~\ref{tab:accu_add_mnist}) when the noise power is from low to medium (e.g,. accuracy decreases less than $0.2\%$ when $\sigma_{\textrm{val},\times}\leq 0.4$ and less than $0.3\%$ when $\sigma_{\textrm{val},+}\leq 0.6$). Note that $\sigma_{\textrm{val},\times}$ is the ratio between standard deviations of noise and magnitude of signals and tolerating $\sigma_{\textrm{val},\times}=0.4$ (SNR = 7.96 dB) is already a good relaxation on the requirement for neuromorphic circuits.

3. If the inference noise power is unknown, noise-injected training with a specified $\sigma_{\textrm{train}}$ also provides consistent accuracy improvement for low-to-medium values of inference noise power (Figure~\ref{fig:cnn6_mnist_mul_add_noisy_training}). For example, the accuracy decreases less than $0.2\%$ when $\sigma_{\textrm{train}}=0.4$ for all inference noise level $\sigma_{\textrm{val}}\leq 0.4$ in both multiplicative and additive noise models.

\begin{figure}[htbp]
	\centering
	\includegraphics[width=1\linewidth]{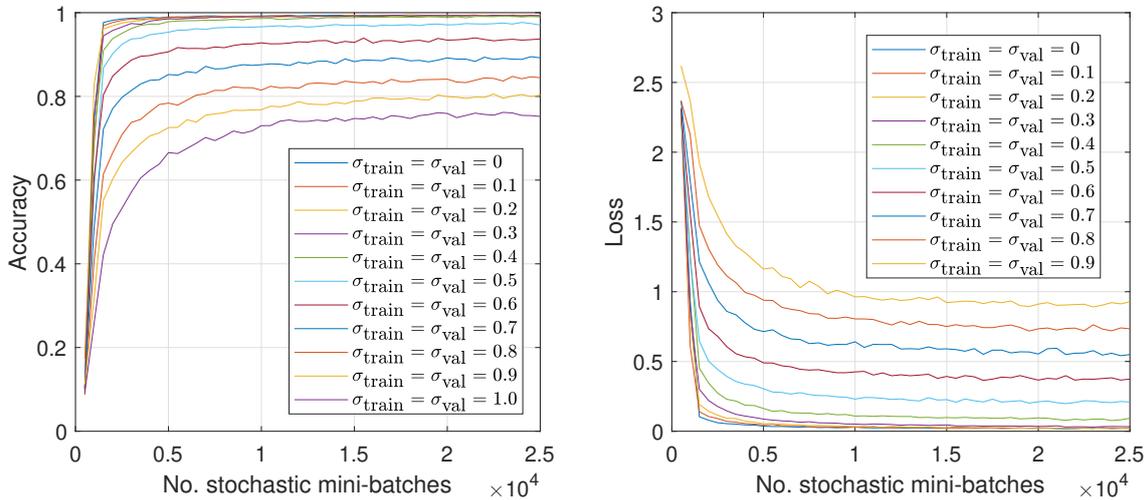}
	\caption{Learning curves of the noise injected training for $\sigma_{\textrm{train},\times}=0.0$ to $1.0$.}
	\label{fig:cnn6_mnist_learning_curve}
\end{figure}

We further investigate the learning speed and the weights of noise injected training. The result for multiplicative noise models is presented in Figure~\ref{fig:cnn6_mnist_learning_curve}, where the learning curve (validation accuracy and loss) for the first $25000$ stochastic mini-batches of size $32$ is shown. It can be observed that higher noise power in training generally results in a lower training speed (with learning rate is fixed to be $0.0001$). 
Table~\ref{tab:l2_weights} shows the expected $L_2$ norm of weights for 11 trained CNN models with different $\sigma_{\textrm{train},\times}$. It can be seen that the magnitude of weights for $\sigma_{\textrm{train},\times}>0$ are similar and are slightly greater than noiselessly trained model.

\begin{table}[htbp]
		\center		
		\caption{Expected $L_2$ norm of all weights of 11 models}
		\label{tab:l2_weights}
		{
			\begin{tabular}{|c|c|c|c|c|c|c|}
				\hline $\sigma_{\textrm{train},\times}$ & 0 & 0.1 & 0.2 & 0.3 & 0.4 & 0.5 \\ 
				\hline $\sqrt{E(w^2)}$	& 0.049	& 0.054	& 0.054	& 0.059	& 0.060	& 0.060	\\
				\hline \hline $\sigma_{\textrm{train},\times}$ &  0.6 & 0.7 & 0.8 & 0.9 & 1.0 & \\ 
				\hline $\sqrt{E(w^2)}$	&  0.058	& 0.056	& 0.060	&0.058 &	0.059 & \\
				\hline 
			\end{tabular} 
		}	
	
\end{table}

\begin{figure}[htbp]
	\centering
	\includegraphics[width=1\linewidth]{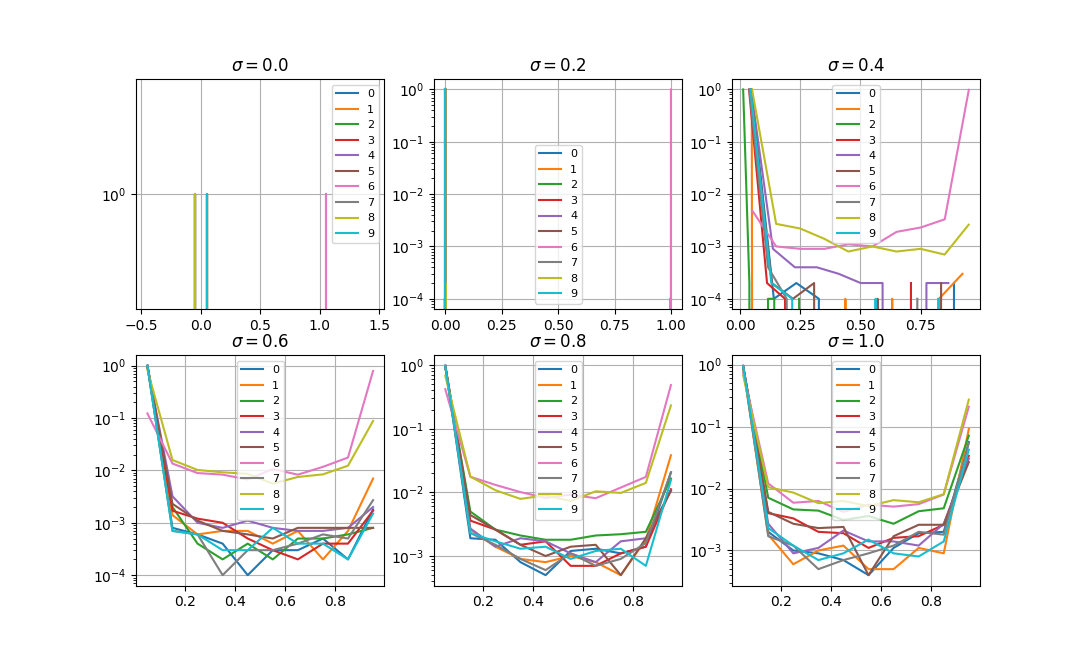}
	\caption{Distribution of the 10 softmax output for MNIST. Model based on conventional training at $\sigma_{\textrm{train},\times}=0.0$, inference with validation noise $\sigma_{\textrm{val},\times}=0$ to $1.0$ at a step of $0.2$.} \vspace{-1em}
	\label{fig:noisyinferenceprobdisttr00}
\end{figure}

\begin{figure}[htbp]
	\centering
	\includegraphics[width=1\linewidth]{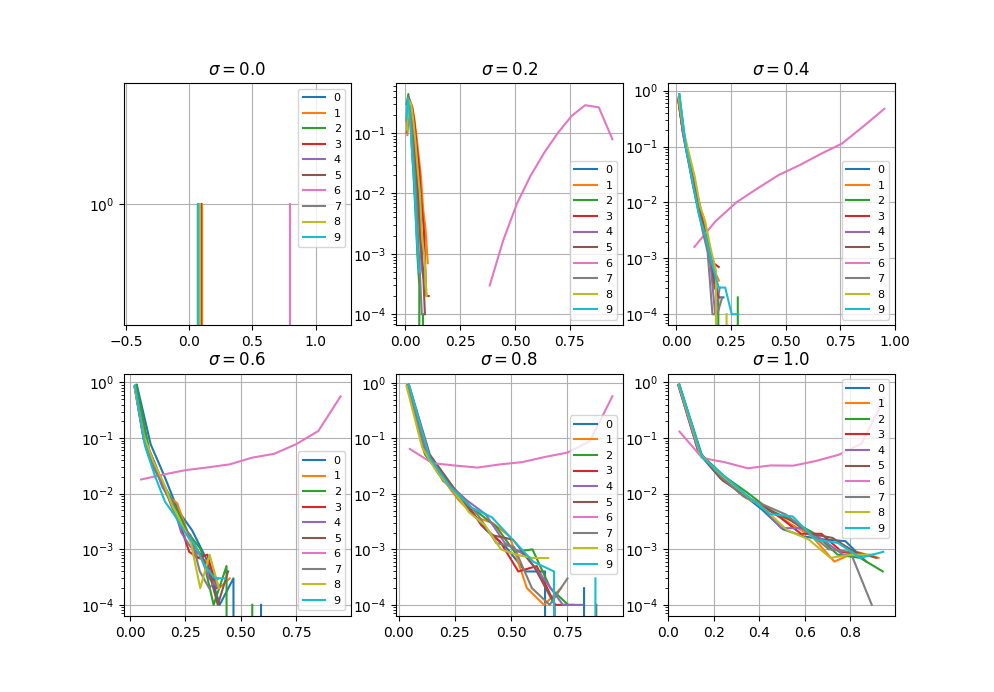}
	\caption{Distribution of the 10 softmax output for MNIST. Model based on noise-injected training at $\sigma_{\textrm{train},\times}=1.0$, inference with validation noise $\sigma_{\textrm{val},\times}=0$ to $1.0$ at a step of $0.2$.}\vspace{-1em}
	\label{fig:noisyinferenceprobdisttr10}
\end{figure}

In order to understand the difference between models from noiseless and noise-injected training, we choose two 6-layer CNN models, corresponding to $\sigma_{\textrm{train},\times}=0.0$ and $1.0$, and run multiplicative noisy inference with $\sigma_{\textrm{val},\times}=0.0$ to $1.0$ for $10000$ times for one same image (which has label ``6'') and then plot the distributions of 10 outputs after the softmax layers, which can be thought as the probability of that image being classified as the corresponding digits. 
Figure~\ref{fig:noisyinferenceprobdisttr00} and Figure~\ref{fig:noisyinferenceprobdisttr10} shows the evolution of the output distributions with increasing $\sigma_{\textrm{val},\times}$. 
Note that the upper right part in each of the six plots indicates higher probability of having large output values. Therefore, it is likely for the neural network to predict the image as labels/curves that appear in that area.
It can be seen that noiselessly trained model (Figure~\ref{fig:noisyinferenceprobdisttr00}) starts to be confused with digit ``6'' (pink) and ``8'' (dark yellow) when $\sigma_{\textrm{val},\times}\geq 0.6$; on the other hand, noise-injected training constantly favors the prediction of ``6''.

\subsubsection{A Depth-40 DenseNet for CIFAR10}
\begin{figure}[htbp]
	\centering
	\includegraphics[width=1\linewidth]{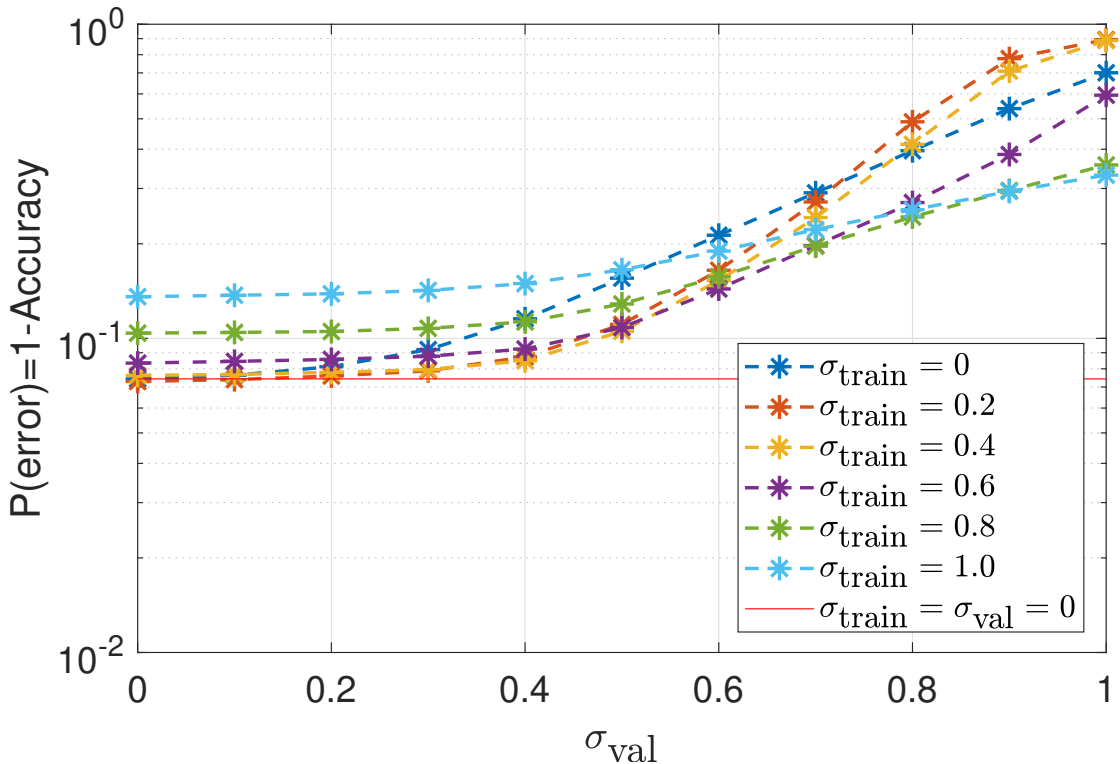}
	\caption{Validation accuracy of the depth-40 DenseNet with Gaussian noise for different $(\sigma_{\textrm{train},\times},\sigma_{\textrm{val},\times})$ pairs.}
	\label{fig:densenet_cifar10_multiplicative_noisy_training}
\end{figure}

We train the non-augmented dataset of CIFAR10 for 300 epochs using momentum optimizer with mini-batch of size 32. We inject multiplicative Gaussian noise right after each matrix-vector multiplication.
Figure~\ref{fig:densenet_cifar10_multiplicative_noisy_training} shows the average accuracy for pairs of $(\sigma_{\textrm{train},\times},\sigma_{\textrm{val},\times})$, where Table~\ref{tab:accu_mul_cifar10} summarizes the results. 

Similar to the 6-layer CNN for MNIST images, we can also conclude from Table~\ref{tab:accu_mul_cifar10} and Figure~\ref{fig:densenet_cifar10_multiplicative_noisy_training} for the multiplicative noise models that noise-injected training provides large robustness gain against noisy inference, e.g., accuracy decreases less than $1.3\%$ (see Table~\ref{tab:accu_mul_cifar10}) if $\sigma_{\textrm{val},\times}\leq 0.4$ (SNR = 7.96 dB).

\begin{table}[htbp]
	\center
	\caption{Accuracy of noisy inference for multiplicative noise models for CIFAR10.}
	\label{tab:accu_mul_cifar10}
	{
		\begin{tabular}{|c|c|c|c|c|c|c|}
			\hline 
			{	\scriptsize$\sigma_{\textrm{val},\times}$} & 0.0 & 0.2 & 0.4 & 0.6 & 0.8 & 1.0 \\ 
			\hline 
			{\scriptsize$\sigma_{\textrm{train},\times}$}=0 & 92.5\% & 91.9\% & 88.5\% & 78.7\% & 60.4\% & 29.9\% \\  
			\hline 
			Best & 92.8\% & 92.4\% & 91.5\% & 85.6\% & 75.7\% & 66.9\% \\ 
			\hline 
			By  {\scriptsize$\sigma_{\textrm{train},\times}$} & 0.1 & 0.2 & 0.4 & 0.6 & 0.8 & 1.0 \\ 
			\hline 
		\end{tabular}
	}
	
\end{table}

%

\subsubsection{LSTMs for MNIST stroke sequences}
\begin{figure}[htbp]
	\centering
	\includegraphics[width=1\linewidth]{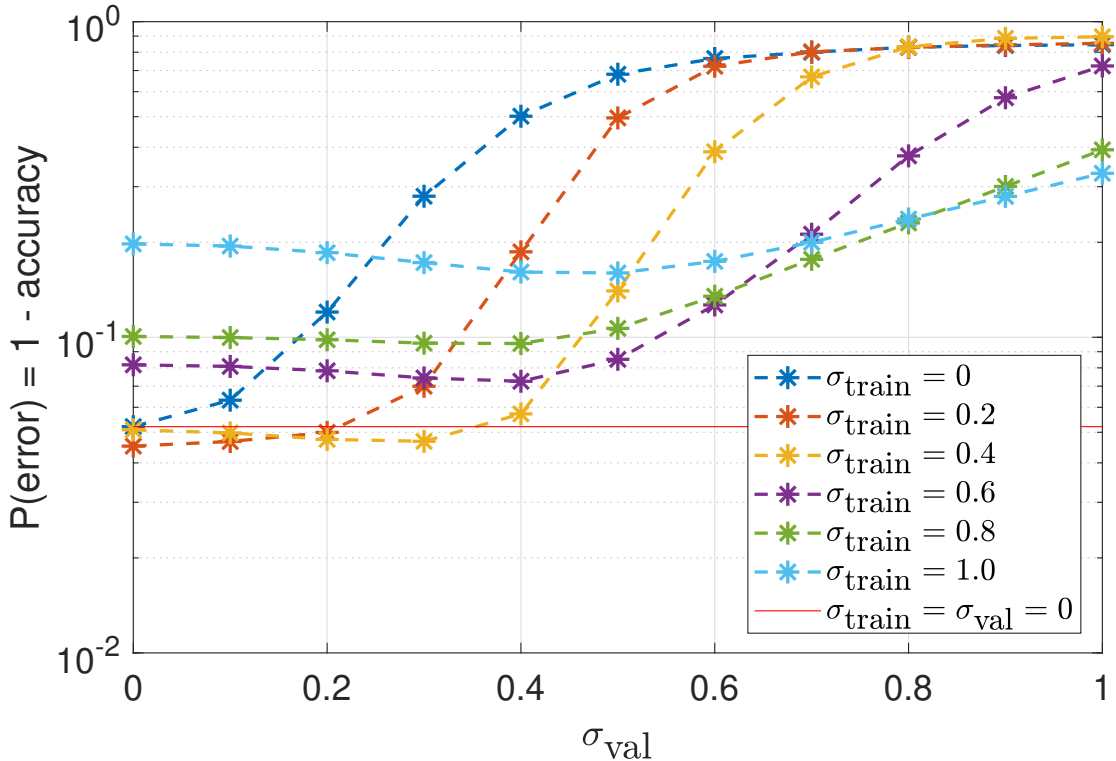}
	\caption{Validation accuracy of LSTM-based RNNs for different $(\sigma_{\textrm{train},\times},\sigma_{\textrm{val},\times})$ pairs.}
	\label{fig:lstmmniststrokemultiplicativenoisytraining}
\end{figure}

This section provides results on classifying stroke sequences on MNIST. We train LSTM-based RNNs for 100 epochs with stochastic mini-batch of batchsize 32 with Adam optimizer, where multiplicative noise are injected after four matrix-vector multiplications within the LSTM cells and one for the final output layer. Figure~\ref{fig:lstmmniststrokemultiplicativenoisytraining} shows the accuracy for different $(\sigma_{\textrm{train},\times},\sigma_{\textrm{val},\times})$ pairs and Table~\ref{tab:accu_mul_mnist_stroke} summarizes the results. The accuracy for noise-injected training decreases less than $1.3\%$ when $\sigma_{\textrm{val},\times}\leq 0.4$, providing much larger robustness than noiseless training against noisy inference.

\begin{table}[htbp]
	\center
	\caption{Accuracy of noisy inference for multiplicative noise models for MNIST stroke sequences.}
	\label{tab:accu_mul_mnist_stroke}
	{
		\begin{tabular}{|c|c|c|c|c|c|c|}
			\hline 
			{	\scriptsize$\sigma_{\textrm{val},\times}$} & 0.0 & 0.2 & 0.4 & 0.6 & 0.8 & 1.0 \\ 
			\hline 
			{\scriptsize$\sigma_{\textrm{train},\times}$}=0 & 94.8\% & 88.0\% & 49.8\% &  23.6\% & 17.0\% & 15.5\% \\  
			\hline 
			Best accu. & 95.6\% & 95.7\% & 94.3\% & 87.3\% & 77.0\% & 66.9\% \\ 
			\hline 
			By  {\scriptsize$\sigma_{\textrm{train},\times}$} & 0.3 & 0.3 & 0.4 & 0.6 & 0.8 &  1.0 \\ 
			\hline 
		\end{tabular}\vspace{-1em}
	}	
\end{table}
%

\subsection{Noisy Inference with Voting}

\begin{figure}
	\centering
	\includegraphics[width=1\linewidth]{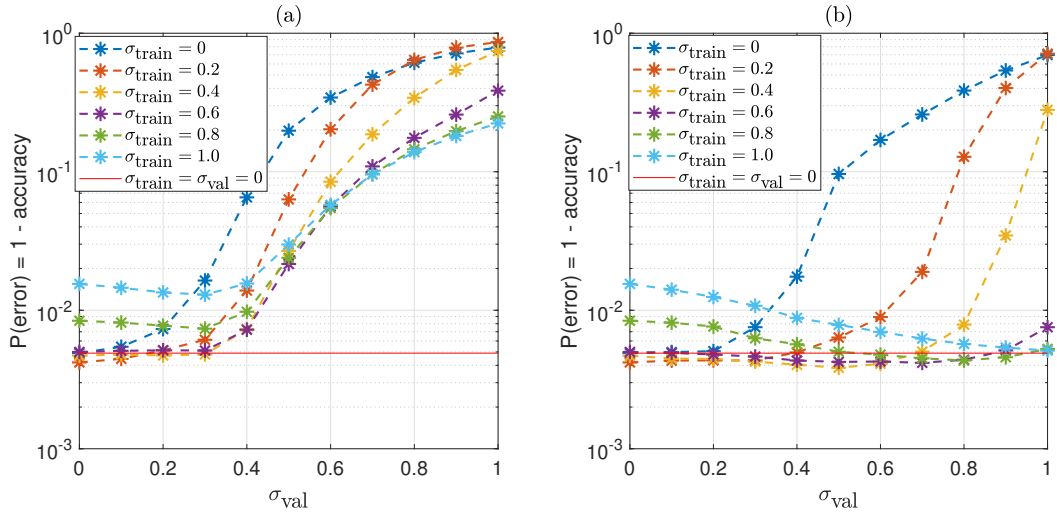}
	\caption{Validation accuracy of the 6-layer CNN for MNIST with multiplicative Gaussian noise of different $(\sigma_{\textrm{train},\times},\sigma_{\textrm{val},\times})$ pairs. (a) One single inference. (b) Inference by voting with 20 runs. }
	\label{fig:cnn6mnistmultiplicativenoisytrainingw-wovotingcomparison}
\end{figure}

\begin{figure}
	\centering
	\includegraphics[width=1\linewidth]{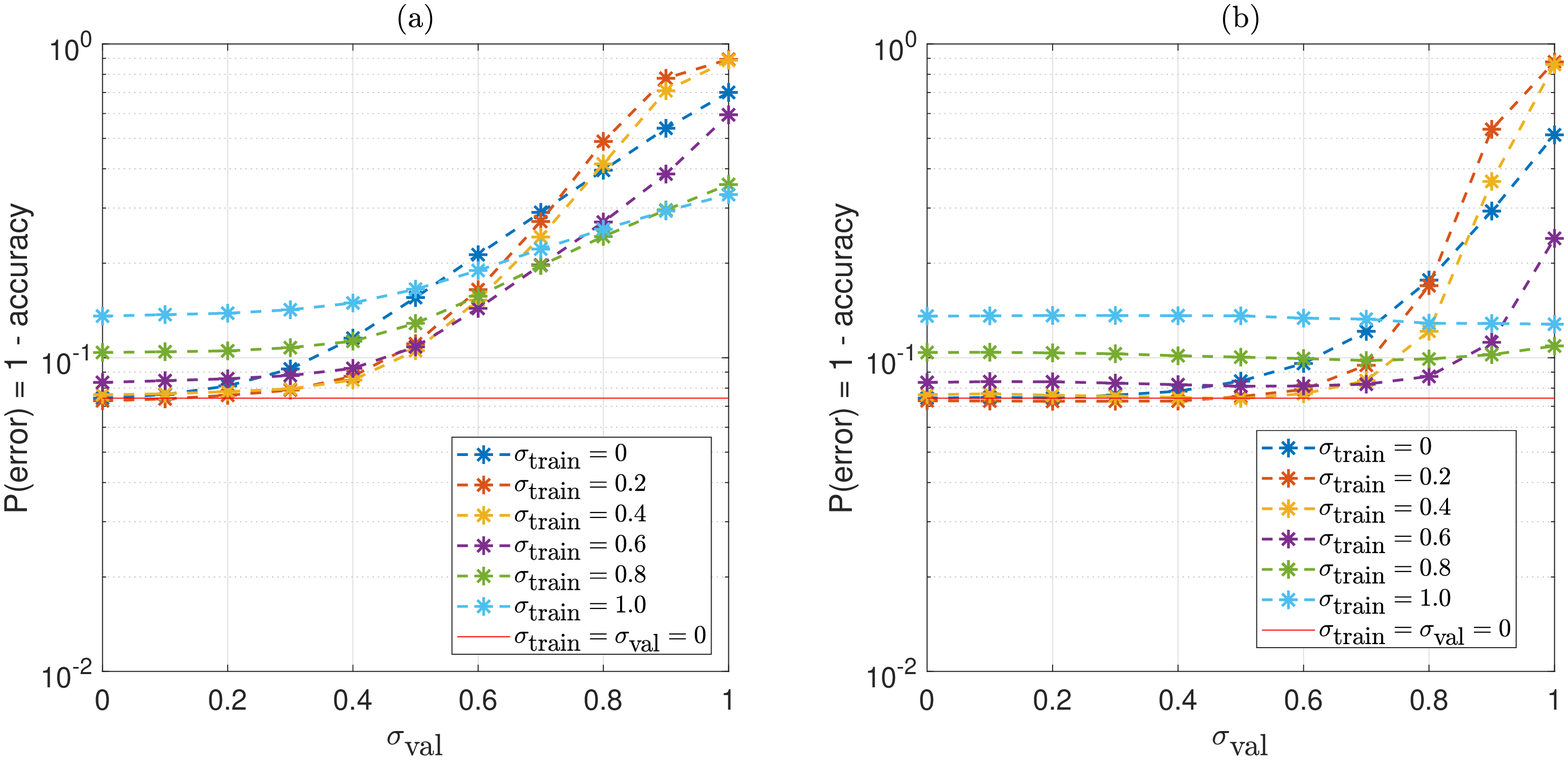}
	\caption{Validation accuracy of the depth-40 DenseNet for CIFAR10 with multiplicative Gaussian noise of different $(\sigma_{\textrm{train},\times},\sigma_{\textrm{val},\times})$ pairs. (a) One single inference. (b) Inference by voting with 20 runs.}
	\label{fig:densenet40cifar10multiplicativenoisytrainingw-wovotingcomparison}
\end{figure}

\begin{figure}
	\centering
	\includegraphics[width=1\linewidth]{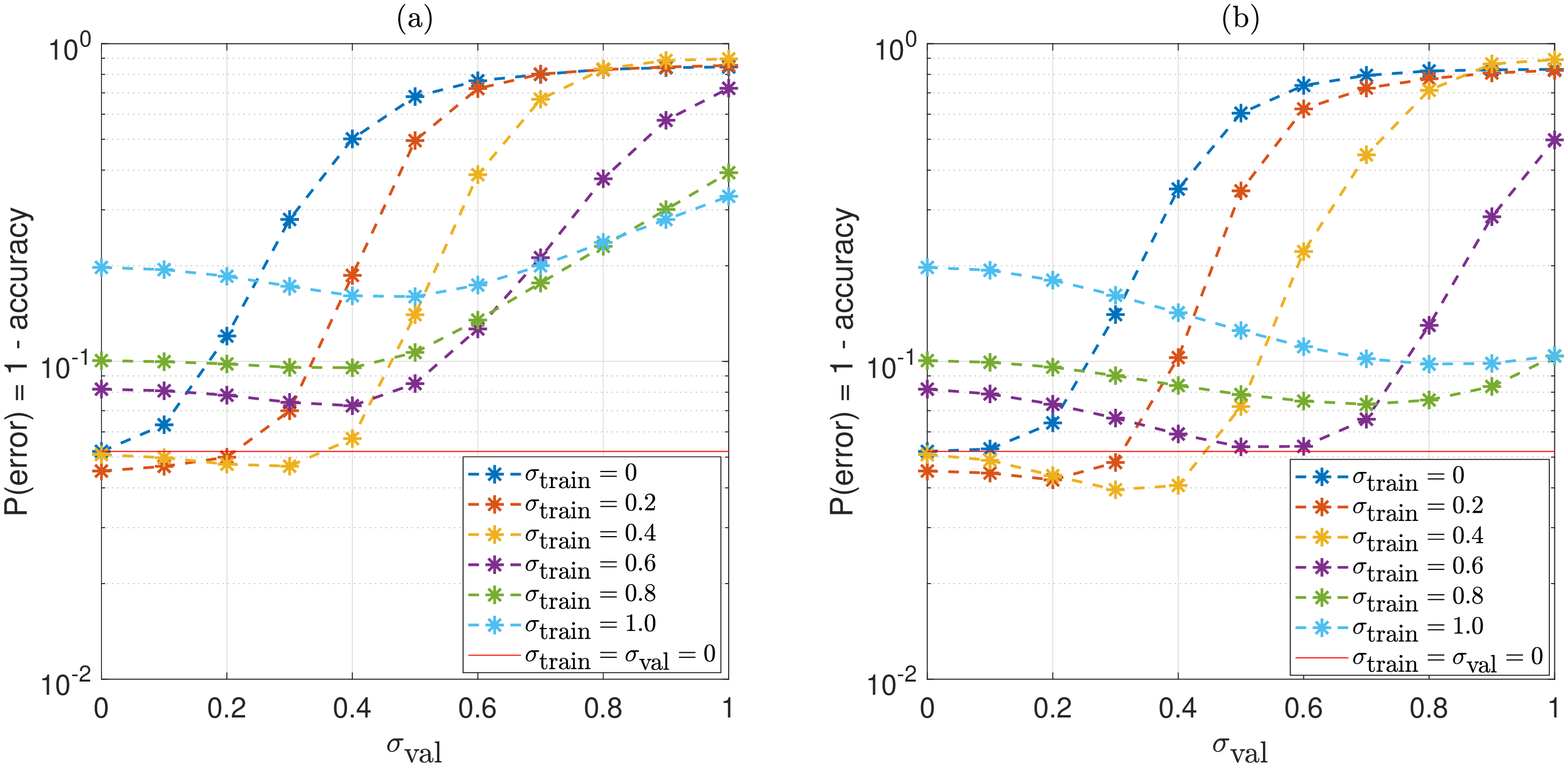}
	\caption{Validation accuracy of the LSTM-based RNN for MNIST stroke sequences with multiplicative Gaussian noise of different $(\sigma_{\textrm{train},\times},\sigma_{\textrm{val},\times})$ pairs. (a) One single inference. (b) Inference by voting with 20 runs.}
	\label{fig:lstmmniststrokemultiplicativenoisytrainingw-wovotingcomparison}
\end{figure}

Inspired by the comparison of Figure~\ref{fig:noisyinferenceprobdisttr00} and Figure~\ref{fig:noisyinferenceprobdisttr10}, we find many instances where the correct prediction is favored in the histogram however not correctly predicted when $\sigma_{\textrm{val}}$ is large, e.g., $1.0$. The reason is that the prediction is probabilistic, i.e., if there is a small overlap in the histogram, e.g., $10\%$, between the favored correct label and a wrong label, the accuracy will be reduces by around $10\%$. 
Observing that the correct prediction appears more often in multiple runs of independent noisy inference, we propose a voting mechanism that can magnify the favored label based on Law of Large Numbers, where we collect a number of predictions of noisy inference and claim the prediction with the most votes (also known as the \textbf{mode}) as the final prediction. Figure~\ref{fig:cnn6mnistmultiplicativenoisytrainingw-wovotingcomparison}, Figure~\ref{fig:densenet40cifar10multiplicativenoisytrainingw-wovotingcomparison}, and Figure~\ref{fig:lstmmniststrokemultiplicativenoisytrainingw-wovotingcomparison} show a detailed comparison for inference with voting (b) and without voting (a), where three (a) sub-figures are replots of Figure~\ref{fig:cnn6_mnist_mul_add_noisy_training}(a), Figure~\ref{fig:densenet_cifar10_multiplicative_noisy_training}, and Figure~\ref{fig:lstmmniststrokemultiplicativenoisytraining} for comparing purposes. Table~\ref{tab:accu_mul_voting_mnist}, Table~\ref{tab:accu_mul_voting_cifar10}, and Table~\ref{tab:accu_mul_voting_mnist_stroke} summarize the results, where the second row repeats the best accuracy for noise-injected training and the third row are the best accuracy achieved by 20 votes on noisy inference using neural network models trained with $\sigma_{\textrm{train},\times}$ equal to the corresponding column in the fourth row.

We can make a few observations based on Figure~\ref{fig:cnn6mnistmultiplicativenoisytrainingw-wovotingcomparison}, Figure~\ref{fig:densenet40cifar10multiplicativenoisytrainingw-wovotingcomparison}, Figure~\ref{fig:lstmmniststrokemultiplicativenoisytrainingw-wovotingcomparison}, Table~\ref{tab:accu_mul_voting_mnist}, Table~\ref{tab:accu_mul_voting_cifar10}, and Table~\ref{tab:accu_mul_voting_mnist_stroke}.

1. While the accuracy in low-to-medium $\sigma_{\textrm{val},\times}$ (i.e., high SNR) regime is similar, voting further improves the accuracy of all three neural networks for three datasets when the noise power is large ($\sigma_{\textrm{val},\times}$ close to $1.0$). For example, when $\sigma_{\textrm{val},\times}=1.0$ (SNR = 0 dB), accuracy is improved by more than $20\%$, from $(77.7\%, 66.9\%, 66.9\%) $ to $(99.5\%, 89.1\%, 89.6\%)$ for MNIST images, CIFAR10, and MNIST stroke sequences, respectively. Compared horizontally, accuracy maintains almost the same ($99.6\%$ vs $99.5\%$)  for MNIST images when $\sigma_{\textrm{val},\times}=1.0$ (SNR=1 in value or 0 dB). Noise-injected training without voting has such high accuracy only when $\sigma_{\textrm{val},\times}\leq 0.2$ (SNR=25 in value or 14 dB), where a 14 dB gain is realized by voting. Similarly, the tolerable $\sigma_{\textrm{val},\times}$ increases from $0.4$ to $0.8$ (a 6 dB SNR gain) for DenseNet of CIFAR10 if $1.5\%$ accuracy loss is acceptable; the tolerable $\sigma_{\textrm{val},\times}$ increases from $0.4$ to $0.6$ (a 3.5 dB SNR gain) for LSTMs of MNIST stroke sequences if $1\%$ accuracy loss is acceptable. Such SNR gains further relax the requirements for neuromorphic circuit designs and enable them to have competitive accuracy with GPU/CPU centered digital computations.

2. We observe in Table~\ref{tab:accu_mul_mnist_stroke} that with voting and noise-injected training combined, noisy inference improves $0.5\%$ accuracy upon noiseless inference for LSTM-based recurrent neural networks, which is counterintuitive since it is often believed that noise during inference would be harmful rather than helpful to the performance. The improvement by noisy inference is also confirmed by independently training multiple LSTM models.

\begin{table}[htbp]
	\center
	\caption{Accuracy of noisy inference by voting for multiplicative noise models for MNIST images.}
	\label{tab:accu_mul_voting_mnist}
	{
		\begin{tabular}{|c|c|c|c|c|c|c|}
			\hline 
			{	\scriptsize$\sigma_{\textrm{val},\times}$} & 0.0 & 0.2 & 0.4 & 0.6 & 0.8 & 1.0 \\ 
			\hline 
			No voting & 99.6\% & 99.5\% & 99.3\% & 94.6\% & 86.1\% & 77.7\% \\ 
			\hline 
			Best voting & 99.6\% & 99.6\% & 99.6\% & 99.6\% & 99.6\% & 99.5\% \\ 
			\hline
			By  {\scriptsize$\sigma_{\textrm{train},\times}$} & 0.2 & 0.5 & 0.5 & 0.5 & 0.8 &  0.9\\ 
			\hline
		\end{tabular}
	}	
\end{table}

\begin{table}[htbp]
	\center
	\caption{Accuracy of noisy inference by voting for multiplicative noise models for CIFAR10.}
	\label{tab:accu_mul_voting_cifar10}
	{
		\begin{tabular}{|c|c|c|c|c|c|c|}
			\hline 
			{	\scriptsize$\sigma_{\textrm{val},\times}$} & 0.0 & 0.2 & 0.4 & 0.6 & 0.8 & 1.0 \\ 
			\hline 
			No voting & 92.8\% & 92.4\% & 91.5\% & 85.6\% & 75.7\% & 66.9\% \\ 
			\hline 
			Best voting & 92.8\% & 92.7\% & 92.7\% & 92.8\% & 91.3\% & 89.1\% \\ 
			\hline
			By  {\scriptsize$\sigma_{\textrm{train},\times}$} & 0.1 & 0.1 & 0.2 & 0.3 & 0.6 &  0.8\\ 
			\hline
		\end{tabular}
		
	}	
\end{table}

\begin{table}[htbp]
	\center
	\caption{Accuracy of noisy inference by voting for multiplicative noise models for MNIST stroke sequences.}
	\label{tab:accu_mul_voting_mnist_stroke}
	{
		\begin{tabular}{|c|c|c|c|c|c|c|}
			\hline 
			{	\scriptsize$\sigma_{\textrm{val},\times}$} & 0.0 & 0.2 & 0.4 & 0.6 & 0.8 & 1.0 \\ 
			\hline 
			No voting & 95.6\% & 95.7\% & 94.3\% & 87.3\% & 77.0\% & 66.9\% \\ 
			\hline 
			Best voting & 95.6\% & \textbf{96.1\%} & 95.9\% & 94.6\% & 92.7\% & 89.6\% \\ 
			\hline
			By  {\scriptsize$\sigma_{\textrm{train},\times}$} & 0.3 & 0.3 & 0.4 & 0.6 & 0.7 &  0.8\\ 
			\hline
		\end{tabular}
		
	}	
\end{table}

\begin{figure}[htbp]
	\centering
	\includegraphics[width=1\linewidth]{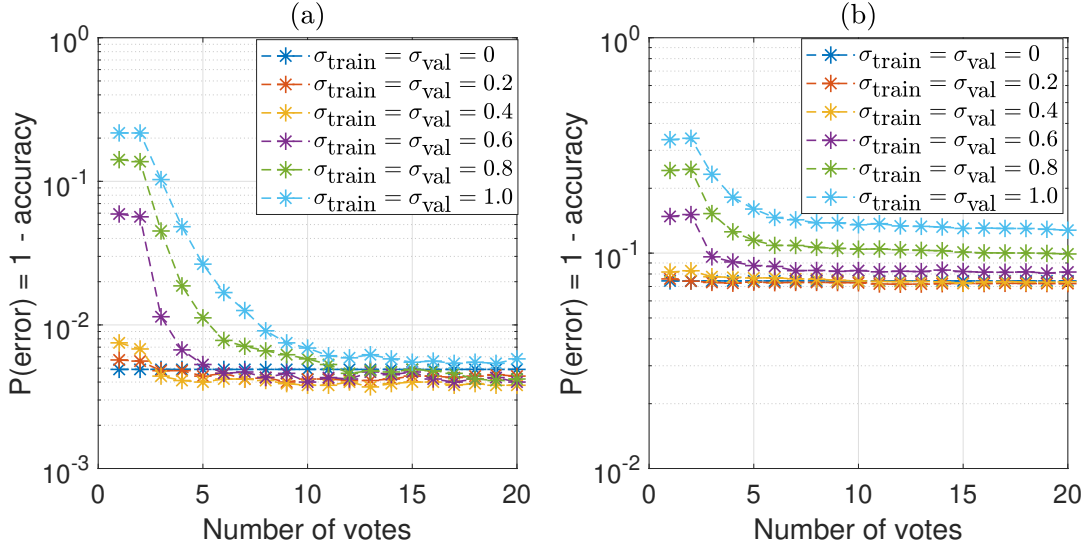}
	\caption{Trade-offs between number of votes and the accuracy of noisy inference for MNIST images (a) and CIFAR10 (b).}
	\label{fig:cnn6mnistmulnoisevotesvsaccu}
\end{figure}

Having observed the benefits of the voting mechanism, we then discuss how to mitigate their weakness in practical applications since latency and power consumption will be increased when a prediction is made by multiple runs of noisy inference. Fig.~\ref{fig:cnn6mnistmulnoisevotesvsaccu}$(a)$ and $(b)$ shows the trade-offs between the number of votes and accuracy for MNIST images and CIFAR10, respectively. It can be seen that the accuracy improves fast when the number of votes increases from $1$ to $10$, in particular for large noise power. Beyond $10$ votes, the accuracy curve becomes almost flat. Fast convergence of the voting is particularly critical for  neuromorphic circuits-based deep neural networks to have smaller latency and power overhead.

The latency overhead can be reduced in two ways. If the chip area is not a bottleneck, then multiple inference can be made in parallel from multiple copies of neuromorphic circuits-based neural networks, resulting in zero latency overhead. Otherwise, it can also be minimized by pipelining multiple inference on a single neuromorphic circuits-based neural network. Each layer of a neural network can be implemented as one memory array and instead of waiting for one inference to finish before the next one to start, images can be fed into the neuromorphic computing system at each time stamp and all layers form a pipeline.
Assume the number of layers is $L$, each layer contributes latency $T$, and the number of votes is $K$, then the overall latency with pipelining is $(L+K)T$ (instead of $KLT$ without pipelining). Compared to the latency of $LT$ for a single inference, the overhead $\frac{(L+K)T - LT}{LT}=\frac{K}{L}$ is slightly greater than $0$ for large $L$, i.e., deeper neural networks, which is becoming widely used in practice.

If a single inference costs power $P$, then the power consumption of $K$ votes would be $KP$, which cannot be reduced by parallel computing or pipelining. However, $P\propto V^2$, where $V$ is the memory programming/sensing voltage, and $V$ is positively related to SNRs, which is the inverse of $\sigma^2_{\textrm{val}}$.  Therefore, smaller $V$ brings with it smaller power consumption at the cost of increasing $\sigma_{\textrm{val}}$. The trade-offs between power consumption and accuracy, connected by $V$ and $\sigma_{\textrm{val}}$ is complicated and beyond the scope of this paper.

\section{Noisy Inference as a Defensive Mechanism for Black-box Adversarial Attacks}

We show in this section that noisy inference can enhance adversarially trained neural networks against black-box adversarial attacks. 
Noisy inference provides stochastic gradients which has been shown in~\cite{CarliniObfuscated18} to provide a false sense of security against white-box attacks. 
However, we observe that the transferability of adversarial examples from one neural network to the other has been weaker if the neural networks used for prediction have noisy inference. We focus on $L_\infty$ attacks where all pixel values of an adversarial example are within an $\epsilon$-cube of a natural one. Our experiments are setup as follows.


1. We use two datasets, MNIST images and CIFAR10 where all pixel values are normalized between $0$ and $1$.

2. The adversarial attacks are iterative fast gradient sign methods (FGSM) with projected gradient descent (PGD) and multiple restarts from a random image within the $\epsilon$-cube of a natural image. It is shown to be a strongest attack based on first-order statistics~\cite{madry2018towards}. The detailed  parameters of the attacks are listed in Table~\ref{tab:adv_para}.
\begin{table}[htbp] 
	\centering
	{
		\begin{tabular}{|c|c|c|}
			\hline 
			& MNIST & CIFAR10 \\ 
			\hline 
			$\epsilon$ & 0.3 & $\frac{8}{256}$ \\ 
			\hline 
			Adv. learn. rate & 0.05 & $\frac{2}{256}$ \\ 
			\hline 
			Iterations & 20 & 7 \\ 
			\hline 
			No. restarts & 20 & 10 \\ 
			\hline 
		\end{tabular} 
		\caption{Parameters of adversarial attacks (iterative FGSM).}
		\label{tab:adv_para}
	}
\end{table}

3. Two neural networks are noiselessly $(\sigma_{\textrm{train}}=0.0)$  and adversarially trained~\cite{madry2018towards}, where each minibatch of size 32 consists of equal number of natural and adversarial samples. One neural network is used to generate adversarial examples (denoted by $\textrm{NN}_{\textrm{adv}}$) and the other is used to validate the accuracy of those adversarial examples (denoted by $\textrm{NN}_{\textrm{val}}$). For MNIST, $\textrm{NN}_{\textrm{adv}}$ is a 4-layer CNN as it is in~\cite{madry2018towards} and $\textrm{NN}_{\textrm{val}}$ is the 6-layer CNN we presented in Table~\ref{tab:cnn6_mnist}. Both neural networks are adversarially trained for 100 epochs. For CIFAR10, $\textrm{NN}_{\textrm{adv}}$ is the depth-40 Densenet with growth rate 12 and $\textrm{NN}_{\textrm{val}}$ is a depth-100 Densenet with growth rate 12. Both neural networks are adversarially trained for 300 epochs.

Tabel~\ref{tab:adv_accu} shows the average validation accuracy for MNIST and CIFAR10 under black-box attacks with 20 runs of noisy inference. It can be observed that the robustness of the adversarially trained neural networks (the 6-layer CNN and the depth-100 Densenet) against the attacks has been further enhanced from $97.40\%$ to $97.90\%$ and from $66.52\%$ to $67.65\%$, respectively. The $0.5\%$ and $1.13\%$ improvements are achieved by multiplicative noisy inference with $\sigma_{\textrm{val},\times}=0.2$ and $0.1$, respectively.

\begin{table}[htbp]
	\centering
	{
		\begin{tabular}{|c|c|c|}
			\hline 
			& MNIST & CIFAR10 \\ 
			\hline 
			$\sigma_{\textrm{val},\times}=0.0$ & 97.40\% & 66.52\%
			\\ 
			\hline 
			$\sigma_{\textrm{val},\times}=0.1$  & 97.79\%	& \textbf{ 67.65\%}
			\\ 
			\hline 
			$\sigma_{\textrm{val},\times}=0.2$ & \textbf{97.90\%} & 67.43\%
			\\ 
			\hline 
			$\sigma_{\textrm{val},\times}=0.3$  & 97.71\% & 66.57\%
			\\ 
			\hline 
			$\sigma_{\textrm{val},\times}=0.4$ & 97.00\%	& 63.39\%
			\\ 
			\hline 
		\end{tabular} 
		\caption{Accuracy of black-box attacks on MNIST and CIFAR10 for different inference noise levels $\sigma_{\textrm{val},\times}$.}
		\label{tab:adv_accu}
	}
\end{table}
\section{Conclusion}
In this paper, we propose to use training with injected noise and inference with voting that imbues neural networks with much greater resilience to imperfect computations during inference, which has potential applications on deep learning with ultra-low power consumptions. 
Three examples of neural network architectures show remarkable improvement on accuracy using these two methods.  With strong noise $(\sigma_{\textrm{val},\times}=1)$, these two methods can improve accuracy from $(21.1\%, 29.9\%, 15.5\%)$ to $(99.5\%, 89.1\%, 89.6\%)$ for the three datasets, respectively. 
With low-to-medium values of noise power ($\sigma_{\textrm{val},\times}\leq 0.4$), noise-injected training itself can improve accuracy from $(95.8\%, 88.5\%, 49.8\%)$ to $(99.4\%,91.5\%,94.3\%)$ for the three datasets at no cost of latency or power consumption.

Further study of black-box attacks against neural networks show $0.5\%$ and $1.13\%$ enhancement for MNIST and CIFAR10, which is brought by noisy inference on adversarially trained neural networks.


\bibliographystyle{elsarticle-num}

\bibliography{reference_minghai}

\end{document}